\documentclass[a4paper]{article}

\usepackage{graphicx}
\usepackage{makecell}
\setcellgapes{2.5pt}
\usepackage{INTERSPEECH2020}
\usepackage{booktabs}
\usepackage{amsmath}
\usepackage{mathtools}
\usepackage{multirow}

\title{Speech to Text Adaptation: Towards an Efficient Cross-Modal Distillation}
\name{Won Ik Cho$^1$, Donghyun Kwak$^2$, Ji Won Yoon$^1$, Nam Soo Kim$^1$}
\address{
  Department of Electrical and Computer Engineering and INMC, Seoul National University$^1$\\
  Search Solution Inc.$^2$}
\email{wicho@hi.snu.ac.kr, donghyun.kwak@navercorp.com, jwyoon@hi.snu.ac.kr, nkim@snu.ac.kr}

\begin{document}

\maketitle
\begin{abstract}


Speech is one of the most effective means of communication and is full of information that helps the transmission of utterer's thoughts. However, mainly due to the cumbersome processing of acoustic features, phoneme or word posterior probability has frequently been discarded in understanding the natural language. Thus, some recent spoken language understanding (SLU) modules have utilized end-to-end structures that preserve the uncertainty information. This further reduces the propagation of speech recognition error and guarantees computational efficiency. We claim that in this process, the speech comprehension can benefit from the inference of massive pre-trained language models (LMs). We transfer the knowledge from a concrete Transformer-based text LM to an SLU module which can face a data shortage, based on recent cross-modal distillation methodologies. We demonstrate the validity of our proposal upon the performance on Fluent Speech Command, an English SLU benchmark. Thereby, we experimentally verify our hypothesis that the knowledge could be shared from the top layer of the LM to a fully speech-based module, in which the abstracted speech is expected to meet the semantic representation.

\end{abstract}

\section{Introduction}

Speech and text are two representative medium of language. Speech, which is delivered mainly via waveform, can be projected to text with the help of automatic speech recognition (ASR). On the contrary, the text is represented visually in letters and is easily digitized to Unicode. It is deemed a lot more beneficial to use text in language comprehension, due to its transmission of information being less uncertain. 

Despite the shared semantic representation between those two \cite{akinnaso1982differences}, especially in engineering studies, they are treated as the data of different modality. In this regard, in contemporary speech-based natural language understanding (NLU) and slot filling tasks, main approaches have exploited either ASR-NLU  pipeline \cite{liu2016attention} or end-to-end speech processing \cite{haghani2018audio,lugosch2019speech,wang2019understanding}. The former, which is conventional, is partially improvable and explainable, while the latter is in fashion since it can mitigate the effect of ASR errors that can be cascaded.

\begin{figure}
	\centering
	\includegraphics[width=\columnwidth]{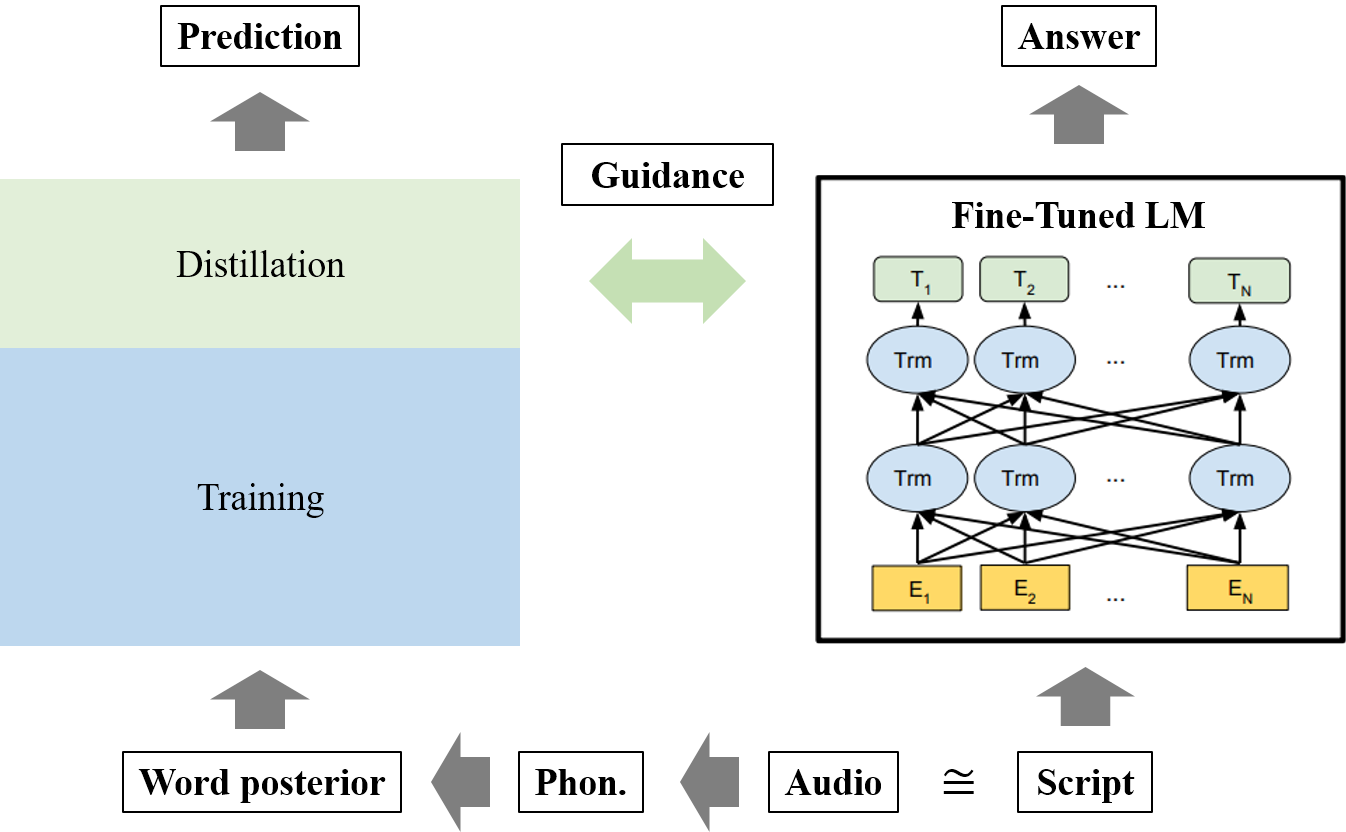}
	\caption{A brief architecture of the proposed distillation scheme on an end-to-end SLU module. The diagram on the right side is adopted from \cite{devlin2018bert}.} \label{fig:fig2}
\end{figure} 

In this paper, we combine the two approaches in a cross-modal viewpoint. Given original speech, its ground truth script, and the target intent, we transfer knowledge from the inference process of the pre-trained language model (LM) to the speech understanding (Figure 1). The core idea is setting a meeting place for the representation from the acoustic data and that from the digitized text, in other words, where the phonetic and lexical data coincide in terms of semantics. In this way, we compensate for the roughness of low-level processing of speech engineering, at the same time benefiting from the text-based inference. The contribution of this study is as the followings:

\begin{itemize}
    \item Leveraging the high-performance inference of text-based fine-tuned LM to an end-to-end spoken language understanding via cross-modal knowledge distillation (KD) 
    \item Verifying the effect of KD with the performance on widely-used intent identification and slot-filling dataset
    \item Suggesting the loss function and KD weight scheduling that can be effective in speech data shortage scenarios
\end{itemize}

\section{Related Work}

Comprehending the directive utterances in terms of intent argument has been vastly investigated so far, whether it be a text or speech input. While the systems with either input aim to execute similar tasks, the speech-based one inevitably requires more delicate handling that owes to the signal-level features.

\subsection{Conventional pipeline}

In conventional settings, spoken language understanding (SLU) is divided into ASR and NLU. ASR is a procedure that transcribes speech into text, and in NLU, the resulting text is analyzed to yield the intent arguments \cite{liu2016attention}. This cascade structure has also been widely used in other spoken language processing tasks, including speech translation \cite{kocabiyikoglu2018augmenting} and intention understanding \cite{cho2019text}. 
It provides a transparent analysis since the modules are distinct, that one can easily recognize the issue and make module-wise enhancement.
However, as \cite{liu2019synchronous} pointed out in the recent study on speech translation, mainly three limitations lie in the pipeline: 1) time delay of cascading, 2) parameter redundancy by module separation, and 3) amplification of ASR errors. Though solutions such as N-best are effective, it is still probable that the last factor induces performance degradation.

\subsection{End-to-end approaches}

To cope with the disadvantages above, in up-to-date SLU, the inference has been performed in an end-to-end manner, wrapping up the ASR and NLU process. Advanced from the early approaches that directly infer the answer from signal level features \cite{qian2017exploring} or jointly trains ASR and NLU components \cite{haghani2018audio}, recent ones use word posterior-level \cite{lugosch2019speech} or phoneme posterior-level \cite{wang2019understanding} pre-trained modules to deal with the shortage of labeled speech resources. The amount of abstraction differs, but the approaches above share the ultimate goal of correctly inferring the argument, usually via slot-wise intent classification. 

\subsection{Pre-trained language models}

Lately, a recurrent neural network (RNN) \cite{schuster1997bidirectional} and Transformer \cite{vaswani2017attention}-based pre-trained LMs \cite{peters2018deep,devlin2018bert} have shown powerful performances over various tasks. Moreover, task-specific training is available by merely adding a shallow trainable layer on the top of the pre-trained module and undertaking a lightweight fine-tuning. However, so far, few end-to-end SLU approaches have taken advantage of them \cite{zhang2019joint} mainly because the inference requires an explicitly text-format input, which necessitates an accurate ASR. Followingly, the task turns into a conventional pipeline problem, deterring the cross-modality. 

\subsection{Knowledge distillation of LMs}

Though the aforementioned limitation is probable, it is a significant loss for the whole SLU inference to renounce the comprehensive and verified information processing of the pre-trained LMs. Is there any approach we can leverage the guaranteed performance? Knowledge distillation (KD) can be one solution \cite{hinton2015distilling}. It is widely used for model compression, but its scheme of minimizing the logit-wise difference can be adopted in the transfer \cite{yim2017gift} or cross-modal \cite{gupta2016cross} learning  as well. Notably for the Transformer \cite{vaswani2017attention}-based pre-trained LMs that occupy a massive volume, recent model compression work proposed the condensation schemes adopting  bidirectional long short-term memory (BiLSTM) \cite{tang2019distilling} or thinner Transformer layers \cite{jiao2019tinybert}. In this paper, we plan to inherit them along with the philosophy of cross-modal distillation.

\section{Proposed Method}			

The core content of our proposal is leveraging the pre-trained LM \cite{devlin2018bert} to SLU via cross-modal fine-tuning, where the tuning is executed in the form of distillation \cite{tang2019distilling,jiao2019tinybert}. 

\subsection{Motivation}

In \cite{akinnaso1982differences}, it is demonstrated in detail how the spoken language and written one share knowledge in abstracting the features. Beyond the lexical features, which are a mere correspondence of a phoneme sequence, written language contains the tonal symbols (e.g., \textit{pinyin}) or punctuation marks, which regard various prosodic features of the speech. Thus, we hypothesized that (1) the integration of both modalities affects a speech-based analysis in a positive way.
 
Consequently, we noted that it had been experimentally displayed that the text-level features reach a state-of-the-art performance within NLU tasks if combined with a pre-trained LM \cite{devlin2018bert}, while yet the speech-oriented models can get little from it. It is not unnatural to expect that (2) the speech processing can be boosted by NLU via some possible form of knowledge sharing. 

In summary, taking into account (1) and (2), we aimed to transfer implicit linguistic processing in LMs (that can help understand the spoken language) to an SLU module, without an explicit process of speech-to-text transformation.

\subsection{Materialization}

The next step is materializing the architecture. Here we refer to two kinds of key papers, namely cross-modal KD for speech translation \cite{liu2019end} and LM compression \cite{tang2019distilling}.

Cross-modal KD is an ambiguous term at a glance since it is difficult to define what the modality is. Thus, we here regard speech and text to incorporate different modality, though in our task, both lead to the same type of inference (intent understanding). Similar to \cite{liu2019end}, where a student speech translation model learns from the prediction of a teacher machine translation module, our SLU model takes advantage of the logit inference of a fine-tuned Transformer-based LM \cite{devlin2018bert}.

In this process, we employ detailed compressing procedures of a Transformer LM \cite{tang2019distilling}, both regarding the model architecture and loss functions. 
At the very first phase, a pre-trained LM, e.g., bidirectional encoder representation from Transformers (BERT) \cite{devlin2018bert}, is fine-tuned with the ground truth, eventually making up a teacher model (though with different modality). 
Consequently, at the end-to-end SLU training phase, which utilizes a frozen pre-trained acoustic module \cite{ravanelli2018speaker,lugosch2019speech}, the loss function is updated with the knowledge distilled from the teacher. Here, knowledge is represented as a loss, which indicates the gap between the logit layers of both modules.

To wrap up, leveraging pre-trained LM to an end-to-end SLU in our approach includes \textit{LM fine tuning} and \textit{distillation from LM to SLU}.

 

\subsection{Model construction}

The final step is constructing the concrete structure of KD, where the teacher pre-trained LM \cite{devlin2018bert} utilizes text input, and the student adopts a speech instance \cite{lugosch2019speech}, while two share the same type of prediction \cite{tang2019distilling}. 
In this process, we set rules of thumb to leverage the given structure and training resources as efficiently as possible. Since one of our aims is to make the best of verified ready-made solutions, we integrated the released structures, the specifications follow as:

\begin{itemize}
    \item Backbone student model adopts ASR pre-trained module \cite{ravanelli2018speaker} and RNN-based intent classifier \cite{lugosch2019speech}, which respectively yields word posterior sequence and slot-wise predictions.
	\item For the teacher model, the pre-trained BERT is utilized without additional modification, and the fine-tuning only exploits a freely available benchmark.
	\item In addition to the cross-entropy (CE) function that is used as the loss of an end-to-end SLU module, a KD loss is augmented to the total loss to reflect the influence of the teacher in the student training phase.
\end{itemize}

In sharing the knowledge, as mentioned above, the guidance is conveyed from the upper components of the fine-tuned BERT logit layers so that the student coincides with the representation that comes from the text input. 
Unlike the raw-text-friendly input layers of LM, we believe that the upper layers are the parts where the abstracted textual information possibly meets the spoken features. 

More specifically, the shared knowledge can be represented as a \textit{regulation} (loss function) that the teacher model gives to the student in the training phase, which leads the tutee to a desirable direction. The notation for the total loss function is as follows:
%
\begin{equation}
L = \alpha_{t} * L_{ce} + \beta_{t} * L_{kd}
\end{equation}
where $t$ is a scheduling factor and $\alpha_{t} + \beta_{t} = 1$. $\alpha_{t}$ and $\beta_{t}$, here denoted as KD weight, are hyper-parameters that decide the influence of $L_{ce}$ and $L_{kd}$ respectively, which can be either fixed or dynamically updated.

Detailed on the losses, $L_{ce}$ is a CE between the answer labels and the predicted logits of the SLU component, as in (2), where $f_{(\cdot)}$ is a logit representation and $Y$ is the target label. $L_{kd}$ is either a mean-squared error (MSE) or smoothed $L_1$ loss (MAE) between the predicted logits of SLU component and the fine-tuned BERT, adopted based on \cite{tang2019distilling} and \cite{chuang2019speechbert} respectively. In (3), $D$ determines the type of distance (e.g., MSE, MAE):
\begin{equation}
L_{ce} = CE(f_{SLU}, Y)
\end{equation}
\begin{equation}
L_{kd} = D(f_{SLU}, f_{BERT})
\end{equation}

In BERT fine-tuning, we adopt two kinds of engineering to investigate the teacher models of diverse performance. For a less accurate one, we build a fully connected (FC) layer on the top of [CLS] representation of BERT \cite{devlin2018bert}, while for the stronger model, we set FC layers for all the output representations of BERT and then apply a max pooling. We call the former \textit{teacher} and the latter \textit{professor} henceforth, considering the difference in training accuracy of both. 

Furthermore, to leverage the \textit{teacher} and \textit{professor} model simultaneously, we mix up the loss that comes from each network to make up a hybrid case as in (4):
\begin{equation}
\begin{array}{l}
L_{kd} = (1 - \gamma) * D(f_{SLU},f_{teacher}) \\ \phantom{L_{kd} .=  }  + \gamma * D(f_{SLU},f_{professor})
\end{array}
\end{equation}
where $\gamma = 0$ denotes \textit{only teacher} and  $\gamma = 1$ \textit{only professor}. For $0 < \gamma < 1$, \textit{hybrid}, we apply the batch-wise intent error rate, $\gamma = err$, inspired be \cite{kim2019knowledge}. This implies that the \textit{professor} models teaches more than \textit{teacher} for the challenging samples.


\section{Experiment}

\subsection{Dataset}

Following the previous end-to-end SLU papers \cite{lugosch2019speech,wang2019understanding,palogiannidi2019end}, we use the Fluent Speech Command (FSC) dataset proposed in \cite{lugosch2019speech}. It incorporates 30,874 English speech utterances annotated with three slots, namely \textit{action}, \textit{object}, and \textit{location}. For example, for ``\textit{Turn the lamp off.}", we have slots filled as \{\textit{action}: decrease, \textit{object}: lamp, \textit{location}: none\}, while  ``\textit{Increase the temperature in the bedroom}'' fills the \textit{location} slot.

We adopt this dataset for three reasons; first, the amount of speakers and speech utterances is substantial, and second, the corpus incorporates fairly complex query-answer pairs; total 248 phrasings with 31 unique intents. Above all, the dataset is publicly available. These qualify the dataset for a benchmark, over other speech command datasets such as 
ATIS \cite{hemphill1990atis}. 
The FSC specification can be found in \cite{lugosch2019speech}. 



\subsection{Implementation}

In our experiment, we referred to three released implementations: (i) a full end-to-end SLU module utilizing FSC\footnote{https://github.com/lorenlugosch/end-to-end-SLU/}, (ii) a freely available pre-trained BERT-Base\footnote{https://github.com/huggingface/transformers}, and (iii) a recipe providing task-specific BERT-to-BiLSTM distillation\footnote{https://github.com/pvgladkov/knowledge-distillation}.
With (i) as a backbone, we distilled the \textit{thinking} of (ii) to the RNN encoder-decoder of (i) in the training phase. The overall procedure follows (iii), which performs a (text-only) BERT-to-BiLSTM distillation and reaches quite a standard (e.g., over ELMo \cite{peters2018deep}). 

\subsubsection{Teacher training and baselines}

Three types of systems are mainly considered. The first type denotes the teacher, namely \textit{pretrained LMs (BERT) fine-tuned with the ground truth (GT) script}, which require an accurate script as an input. Teacher training was done with the whole FSC scripts, tokenized via word piece model \cite{wu2016google} of BERT-Base, maximum length 60. For \textit{teacher} we achieved the train error rate of 3.74\%, and for \textit{professor}, 0.19\%. Both reached the test error rate of 0.00\%. 

For the teachers, if ASR output transcriptions are fed as input, we acquire the systems of the second type; an \textit{ASR-NLU pipeline}, a common baseline. We did not re-train the ASR module with FSC, and instead used recently distributed Jasper \cite{li2019jasper} module 
with high accuracy (LibriSpeech \cite{panayotov2015librispeech} WER 3.61\%)
. It was observed that \textit{teacher} gets test error rate of 18.18\%, while \textit{professor} reaches 16.75\%, showing slightly more robustness. 

The last type of models are speech-based ones: a \textit{word posterior-based RNN end-to-end} \cite{lugosch2019speech} and a \textit{phoneme posterior-based Transformer architectures} \cite{wang2019understanding}. In specific, \cite{lugosch2019speech} exploits the intermediate layer of ASR pretrained model, and besides, \cite{wang2019understanding} trains new BERT \cite{devlin2018bert} or ERNIE \cite{zhang2019ernie}-like networks with the phoneme posterior of the acoustic model component as an input. Both utilize the non-textual representation in a task-specific tuning, and especially \cite{wang2019understanding} performs a large-scale.phone-level pretraining. Unlike \cite{lugosch2019speech}, which we train as well in our environment, for \cite{wang2019understanding},  the reported results are adopted from the original paper, especially the highest among all the settings. For the test of these models, only the speech inputs are utilized.


\subsubsection{The proposed}

We compare the above approaches to the proposed scheme. As stated in Section 3.3, the whole process resembles \cite{lugosch2019speech}, only with the difference in the total loss $L$.
Mainly three factors determine $L_{kd}$: \textit{who teaches}, \textit{how the loss is calculated}, and \textit{how much the guidance influences}. The first one regards the source of distillation, namely \textit{teacher} and \textit{professor}. The second is upon $D$, MSE or MAE. The last denotes the scheduling on $\alpha_{t}$ and $\beta_{t}$. 


On the last topic, where $\alpha_t$ and $\beta_t$ sets the KD weight, we perform three scheduling strategies regarding the temporal factor, namely $t$ the epochs. 
\begin{equation}
\begin{array}{l}
(a) \phantom{(a)} \beta_t = err_{t, batch} (= 1-acc_{t, batch}) \\ (b) \phantom{(a)} \beta_t = exp(1-t) \\ (c) \phantom{(a)} \beta_t = 0.1 * max(0, -|t-\mu|/(0.5 * \mu) + 1)
\end{array}
\end{equation}
First one is the aforementioned $err$, adopted as (5a), which depends upon the training intent error rate per batch. Qualitatively, it regards well-classified samples contribute more to the training, as suggested in \cite{kim2019knowledge}. Second one is the exponential decay (\textit{Exp.}), calculated as (5b) where the teacher influence falls down exponentially and mechanically depending on the epoch. The rest is the triangular scheduling (\textit{Tri.}) which is inspired by \cite{yang2019towards}, defined as (5c) for $\mu = T/2$ and $T$ the maximum number of epochs. $0.1$ was multiplied to compensate for the scale of KD loss compared to CE. Unlike \textit{Exp.} where the teacher warms up the parameters at the very early phase, in \textit{Tri.}, the student learns by itself at first and the teacher intervenes in the middle. 


%
%
%
%
%
%
%
%
%
%
%


\subsection{Result and discussion}

\subsubsection{Teacher performance}

Overall, it is verified that the fine-tuned text models show significance with the ground truth script, albeit \textit{teacher} failed to reach the performance of some end-to-end SLU models in terms of training accuracy. 
The text-based systems seem to face less amount of uncertain representations in the training phase. Besides, \textit{professor} far outperforms \textit{teacher} in training accuracy, as shown in Section 4.2.1. 
However, since the performance is not sustained in the ASR context, we set a baseline for ASR-NLU with the borrowed value \cite{wang2019understanding} (Table 1). 


\subsubsection{Comparison and analysis}

The results show that the distillation affects if the setting is considerate, 1.19\% to 1.02\% at best (Table 1). 
Though some do not display the enhancement probably for the sensitivity of the test set, we obtained the performance regarding phoneme BERT (1.05\%) \cite{wang2019understanding} for certain cases, namely utilizing \textit{teacher} and \textit{hybrid}. Though we could not achieve the current best-known state that adopts ERNIE (0.98\%), one of ours with MAE reached slightly beyond BERT. We acquired around 15\% reduced error rate via simple distillation to the vanilla SLU model.

It is notable that \textit{professor} does not necessarily present the best teaching, in correspondence with the recent findings of \cite{yoon2020tutornet}.  
It was also observed that the \textit{professor} distillation spent much more epochs for the student to reach the fair accuracy in the training phase. In this regard, for \textit{data shortage scenario \#1}, even  \textit{hybrid} (where \textit{professor} influences much) failed to converge, with \textit{err} scheduling that had yielded the best performance. This implies that the distillation should be more like guidance, not just a harsh transfer, if the resource is scarce. 

The decision of loss function is also the part we scrutinized in this study considering the previous research on Speech BERT \cite{chuang2019speechbert}. It has been empirically shown that MAE can compensate for the different natures of the speech and text data. This is not significant in the \textit{whole-data} scenario (Table 1), where the overfitting is less probable. However, in data shortage scenarios, adopting MSE failed to guarantee the usefulness of distillation as a helper, inducing degradation or collapse (Table 2). We assume that this is a matter of the boosted scale of the loss, that comes from the different levels of uncertainty of both modalities, which appears even with MAE sometimes (\textit{scenario \#2}).

\subsubsection{Data shortage scenarios and scheduling}


We checked that the proposed method is also useful in the case where the amount of text data dominates the speech, by restricting the usage of speech-text pairs to 10\% and 1\% in the training phase (Table 2). 
Given the identical test set for all the scenarios, the amount of error reduction became more visible as the data decreased. For instance with \textit{teacher}, MAE, and \textit{Exp.}, we obtained 0.9\%p error rate reduction for \textit{whole-data}, 0.16\% for 10\% scenario, and 0.44\% for 1\% scenario.

Under shortage, the scheduling seems to matter more than the  \textit{whole-data} case. At first we suspected that \textit{err} or \textit{Tri.} would show considerable performance. However, for the both scenarios, exponential decay (\textit{Exp.}) exhibited the significance compared to the others, given MAE and \textit{teacher} distillation. This means that early influence and fading away can lead the student to better direction if the resource is not enough (\textit{Exp.} $>$  \textit{err}, \textit{Tri.}). The teaching should be moderate (\textit{teacher} $>$ \textit{hybrid}), and the transfer of loss should be restricted in some circumstances (e.g., $\beta_t$ = \textit{err} in \textit{scenario \#2}) to prevent the collapse.



\begin{table}[]
	\centering
	\makegapedcells
	\resizebox{0.94\columnwidth}{!}{%
		\begin{tabular}{|l|c|c|c|c|}
			\hline
			\multicolumn{1}{|c|}{\textbf{Test error rate (\%)}}               & \multicolumn{4}{c|}{\textit{\textbf{Reported \& done}}}                                                                                                                                       \\ \hline
			ASR-NLU                                                           & \multicolumn{4}{c|}{16.75 (Done) / \textbf{9.89} (Reported by \cite{wang2019understanding})}                                                                                                                                                                     \\ \hline
			Lugosch et al. \cite{lugosch2019speech}                                                  & \multicolumn{4}{c|}{1.20 / \textbf{1.19} (Done)}                                                                                                                                                       \\ \hline
			Wang et al. \cite{wang2019understanding}                                                      & \multicolumn{4}{c|}{\textbf{1.05} (BERT) / \textbf{0.98} (ERNIE)}                                                                                                                                               \\ \hline
			\multicolumn{1}{|r|}{\multirow{3}{*}{\textit{\textbf{Proposed}}}} & \multirow{3}{*}{\begin{tabular}[c]{@{}c@{}}$\beta_t$  = 0.1\\ MSE\end{tabular}} & \multirow{3}{*}{\begin{tabular}[c]{@{}c@{}}$\beta_t$  = 0.5\\ MSE\end{tabular}} & \multicolumn{2}{c|}{$\beta_t$ =  \textit{err}}            \\ \cline{4-5} 
			\multicolumn{1}{|r|}{}                                            &                                                                        &                                                                        & \multirow{2}{*}{MSE} & \multirow{2}{*}{MAE} \\
			\multicolumn{1}{|r|}{}                                            &                                                                        &                                                                        &                      &                      \\ \hline
			Distill-\textit{Teacher} ($\gamma$ = 0)                                           & 1.19                                                                   & 1.19                                                                   & \textbf{1.05}        & 1.18                 \\ \hline
			Distill-\textit{Professor} ($\gamma$ = 1)                                         & 1.18                                                                   & 1.19                                                                   & 1.13                 & 1.18                 \\ \hline
			Distill-\textit{Hybrid} ($\gamma$ = \textit{err})                                          & 1.13                                                                   & 1.13                                                                   & \textbf{1.05}        & \textbf{1.02}        \\ \hline
		\end{tabular}%
	}
	\caption{Results of the whole-data scenario.}
	\label{tab:my-table}
\end{table}

\begin{table}[]
	\centering
	\makegapedcells
	\resizebox{0.85\columnwidth}{!}{%
		\begin{tabular}{|l|c|c|c|c|}
			\hline
			\multicolumn{1}{|c|}{\multirow{2}{*}{\textbf{Test error rate (\%)}}} & \multirow{2}{*}{\textit{\begin{tabular}[c]{@{}c@{}}\textbf{MSE}\\ (err)\end{tabular}}} & \multicolumn{3}{c|}{\textit{\textbf{MAE + Schedulings}}} \\ \cline{3-5} 
			\multicolumn{1}{|c|}{}                                               &                                                                                        & \textit{err}      & \textit{Exp.}     & \textit{Tri.}    \\ \hline
			Distill-\textit{Teacher} ($\gamma$ = 0)                                              & \textbf{1.05}                                                                          & 1.18              & 1.10              & \textbf{1.05}    \\ \hline
			Distill-\textit{Professor} ($\gamma$ = 1)                                            & 1.13                                                                                   & 1.18              & 1.18              & 1.08             \\ \hline
			Distill-\textit{Hybrid} ($\gamma$ = \textit{err})                                             & \textbf{1.05}                                                                          & \textbf{1.02}     & 1.08              & 1.08             \\ \hline
			\multicolumn{1}{|r|}{\textit{\textbf{Data shortage \#1}}}            & \multicolumn{4}{c|}{\textit{\textbf{10\%} (10 random subsets)}}                                                                                            \\ \hline
			Lugosch et al.  \cite{lugosch2019speech}                                                      & \multicolumn{4}{c|}{2.10 / \textbf{2.04} (Done)}                                                                                                           \\ \hline
			Distill-\textit{Teacher} ($\gamma$ = 0)                                              & 2.32                                                                                   & 2.00              & \textbf{1.88}     & 1.98             \\ \hline
			Distill-\textit{Hybrid} ($\gamma$ = \textit{err})                                             & $\times$                                                                                     & 2.06              & 2.01              & 1.98             \\ \hline
			\multicolumn{1}{|r|}{\textit{\textbf{Data shortage \#2}}}            & \multicolumn{4}{c|}{\textit{\textbf{1\%} (20 random subsets)}}                                                                                             \\ \hline
			Lugosch et al. \cite{lugosch2019speech}                                                       & \multicolumn{4}{c|}{\textbf{17.22} (Done)}                                                                                                                 \\ \hline
			Distill-\textit{Teacher} ($\gamma$ = 0)                                              & $\times$                                                                                       & $\times$                 & \textbf{16.88}    & 17.27            \\ \hline
		\end{tabular}%
	}
	\caption{Distillation influences in the data shortage scenarios with various scheduling schemes. We set \cite{lugosch2019speech} as baseline for the shortage scenarios. $\times$ denotes the failure of convergence.}
	\label{tab:my-table}
\end{table}

\subsubsection{Knowledge sharing}
One may ask whether the distillation is truly a sharing of knowledge, since it can be interpreted as merely supervising the student based on relatively accurate logits. Also, in view of probabilistic distribution, some outputs regarding \textit{confident} inferences might be just considered as the hard-labeled answer itself. 
However, in quite a few cases, logits can reflect the extent each problem is difficult for the teacher. We believe that such information is intertwined with the word-level posterior, which incorporates the uncertainty of speech processing as well.

\section{Conclusion}

In this paper, we materialized the speech to text adaptation by an efficient cross-modal LM distillation on an intent classification and slot filling task, FSC. 
We found that cross-modal distillation works in SLU, and more significantly in speech data shortage scenarios, with a proper weight scheduling and loss function. It also appears that an uncompetitive teacher conveys more useful knowledge.
As future work, we plan to decompose the layer-wise information hierarchy of pre-trained LMs that the SLU systems might leverage beyond logit-level representations. 

\section{Acknowledgements}

This research was supported by NAVER Corp. The authors appreciate Hyoungseok Kim, Gichang Lee, and Woomyoung Park for constructive discussion. Also, the authors greatly thank Sang-Woo Lee, Kyoung Tae Doh, and Jung-Woo Ha for helping this research.



%



\end{document}